\documentclass{article}
\usepackage{authblk}
\usepackage[utf8]{inputenc}
\usepackage{main}
\usepackage{microtype}
\usepackage{subcaption}
\usepackage{graphicx}
\usepackage{times}
\usepackage{latexsym}
\usepackage{amsmath}
\usepackage{float}
\usepackage{footnote}
\usepackage{enumitem}
\usepackage{bm}
\usepackage{arydshln}
\usepackage{booktabs}
\usepackage{multicol}
\usepackage{multirow}
\usepackage{color}
\usepackage{xcolor}     
\usepackage{colortbl}
\usepackage{bbding}
\usepackage{makecell}
\usepackage{mathtools}
\usepackage{imakeidx}
\usepackage{longtable}
\usepackage{wrapfig}
\usepackage{rotating}
\makeindex
\usepackage{arydshln}
\usepackage{lipsum}
\usepackage{natbib}
\usepackage[toc]{multitoc}
\usepackage[edges]{forest}
\usepackage[normalem]{ulem}
\definecolor{mydarkblue}{rgb}{0,0.08,0.45}
\usepackage[colorlinks=true,linkcolor=mydarkblue,citecolor=mydarkblue,filecolor=mydarkblue,urlcolor=mydarkblue]{hyperref}
\usepackage{CJKutf8}
\usepackage{awesomebox} 
\usepackage{bbding}
\usepackage[most]{tcolorbox}
\usepackage{booktabs}
\usepackage{geometry}
\definecolor{wkblue}{rgb}{0.2, 0.3, 0.6}
\definecolor{meta-color}{rgb}{0.5, 0.5, 0.5}
\usepackage{amsmath}
\usepackage{enumitem}
\usepackage{lscape} 
\usepackage{booktabs}
\usepackage{algorithm}    

\usepackage{algorithmicx}
\usepackage{algpseudocode}
\usepackage{tabularx,booktabs}
\usepackage{makecell}

\usepackage{amssymb}
\usepackage{amsfonts}

\usepackage[tikz]{bclogo}
\usepackage[framemethod=tikz]{mdframed}
\definecolor{bgblue}{RGB}{245,243,253}
\definecolor{ttblue}{RGB}{91,194,224}

\mdfdefinestyle{mystyle}{%
  rightline=true,
  innerleftmargin=10,
  innerrightmargin=10,
  outerlinewidth=3pt,
  topline=false,
  rightline=true,
  bottomline=false,
  skipabove=\topsep,
  skipbelow=\topsep
}

\newtcolorbox{myboxi}[1][]{
  breakable,
  title=#1,
  colback=red!5,
  colbacktitle=red!5,
  coltitle=black,
  fonttitle=\bfseries,
  bottomrule=0pt,
  toprule=0pt,
  leftrule=2pt,
  rightrule=2pt,
  titlerule=0pt,
  arc=0pt,
  outer arc=0pt,
  colframe=red,
}

\newtcolorbox{myboxnote}[1][]{
  breakable,
  title=#1,
  colback=orange!0,
  colbacktitle=orange!0,
  coltitle=black,
  fonttitle=\bfseries,
  bottomrule=0pt,
  toprule=0pt,
  leftrule=2pt,
  rightrule=2pt,
  titlerule=0pt,
  arc=0pt,
  outer arc=0pt,
  colframe=orange,
}

\newtcolorbox{myboxii}[1][]{
  breakable,
  freelance,
  title=#1,
  colback=white,
  colbacktitle=white,
  coltitle=black,
  fonttitle=\bfseries,
  bottomrule=0pt,
  boxrule=0pt,
  colframe=white,
  overlay unbroken and first={
  \draw[red!75!black,line width=3pt]
    ([xshift=5pt]frame.north west) -- 
    (frame.north west) -- 
    (frame.south west);
  \draw[red!75!black,line width=3pt]
    ([xshift=-5pt]frame.north east) -- 
    (frame.north east) -- 
    (frame.south east);
  },
  overlay unbroken app={
  \draw[red!75!black,line width=3pt,line cap=rect]
    (frame.south west) -- 
    ([xshift=5pt]frame.south west);
  \draw[red!75!black,line width=3pt,line cap=rect]
    (frame.south east) -- 
    ([xshift=-5pt]frame.south east);
  },
  overlay middle and last={
  \draw[red!75!black,line width=3pt]
    (frame.north west) -- 
    (frame.south west);
  \draw[red!75!black,line width=3pt]
    (frame.north east) -- 
    (frame.south east);
  },
  overlay last app={
  \draw[red!75!black,line width=3pt,line cap=rect]
    (frame.south west) --
    ([xshift=5pt]frame.south west);
  \draw[red!75!black,line width=3pt,line cap=rect]
    (frame.south east) --
    ([xshift=-5pt]frame.south east);
  },
}

\usepackage{fancyhdr} 
\usepackage{blindtext} 
\usepackage{makecell}

\DeclareCaptionFont{black}{\color{black}}

\definecolor{myblue}{rgb}{0.9, 0.1, 0.94}
\definecolor{mygreen}{rgb}{0.64, 0.56, 0.88}
\definecolor{myyellow}{rgb}{0.68, 0.6, 0.1}
\definecolor{fancygreen}{rgb}{0.33, 0.68, 0.20}
\definecolor{salmon}{rgb}{0.94, 0.52, 0.49}
\definecolor{tablegreen}{rgb}{0.82, 0.94, 0.75}
\definecolor{tableblue}{rgb}{0.81, 0.90, 0.94}
\definecolor{tablered}{rgb}{0.97, 0.85, 0.85}
\definecolor{tableorange}{rgb}{0.96, 0.85, 0.81}

\newenvironment{itemize*}%
 {\leftmargini=10pt\begin{itemize}%
  \setlength{\itemsep}{0pt}%
  \setlength{\parskip}{0pt}%
  }%
 {\end{itemize}}
\newenvironment{enumerate*}%
 {\begin{enumerate}%
  \setlength{\itemsep}{0pt}%
  \setlength{\parskip}{0pt}}%
 {\end{enumerate}}

\usepackage{xcolor}
\usepackage{listings}

\newcommand\JSONnumbervaluestyle{\color{blue}}
\newcommand\JSONstringvaluestyle{\color{red}}

\newif\ifcolonfoundonthisline

\makeatletter

\lstdefinestyle{json}
{
  showstringspaces    = false,
  keywords            = {false,true},
  alsoletter          = 0123456789.,
  morestring          = [s]{"}{"},
  stringstyle         = \ifcolonfoundonthisline\JSONstringvaluestyle\fi,
  MoreSelectCharTable =%
    \lst@DefSaveDef{`:}\colon@json{\processColon@json},
  basicstyle          = \ttfamily,
  keywordstyle        = \ttfamily\bfseries,
}

\newcommand\processColon@json{%
  \colon@json%
  \ifnum\lst@mode=\lst@Pmode%
    \global\colonfoundonthislinetrue%
  \fi
}

\lst@AddToHook{Output}{%
  \ifcolonfoundonthisline%
    \ifnum\lst@mode=\lst@Pmode%
      \def\lst@thestyle{\JSONnumbervaluestyle}%
    \fi
  \fi
  \lsthk@DetectKeywords%
}

\lst@AddToHook{EOL}%
  {\global\colonfoundonthislinefalse}

\makeatother

\usepackage{etoolbox}
\usepackage{natbib}
\usepackage{url}
\newcounter{bibcount}
\makeatletter
\patchcmd{\@lbibitem}{\item[}{\item[\hfil\stepcounter{bibcount}{[\thebibcount]}}{}{}
\renewcommand\NAT@bibsetup%
  [1]{\setlength{\leftmargin}{\bibhang}\setlength{\itemindent}{-\parindent}%
      \setlength{\itemsep}{\bibsep}\setlength{\parsep}{\z@}}
\makeatother

\newcommand*\samethanks[1][\value{footnote}]{\footnotemark[#1]}

\definecolor{mybrown}{RGB}{128,64,0}

\definecolor{titlecolor}{HTML}{4c9cff}

\begin{document}


\title{LIMR: Less is More for RL Scaling}

\author{%
\textbf{Xuefeng Li\thanks{~~Co-first authors}\space\space\space 
Haoyang Zou\samethanks\hspace{0.5em}
Pengfei Liu}\thanks{~~Corresponding author}\\
SJTU, SII, GAIR}
  
\maketitle
\thispagestyle{fancy}
\fancyhead{}
\lhead{\includegraphics[height=0.67cm]{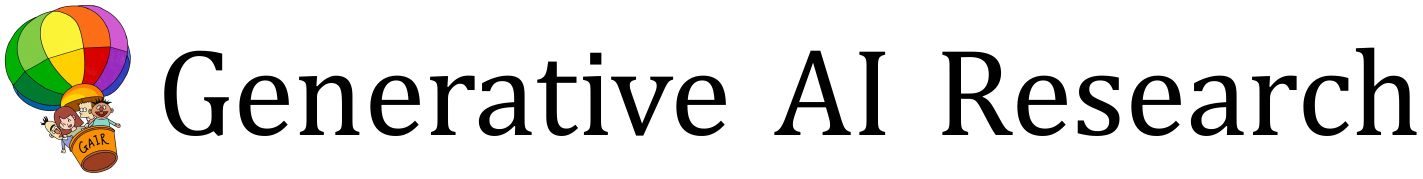}}
\renewcommand{\headrulewidth}{0pt}
\setlength{\headsep}{0mm}

\begin{abstract}



In this paper, we ask: \textbf{what truly determines the effectiveness of RL training data for enhancing language models' reasoning capabilities?} While recent advances like o1, Deepseek R1, and Kimi1.5 demonstrate RL's potential, the lack of transparency about training data requirements has hindered systematic progress.
Starting \textbf{directly from base models without distillation}, we challenge the assumption that scaling up RL training data inherently improves performance.
we demonstrate that a strategically selected subset of just \textbf{1,389 samples can outperform the full 8,523-sample dataset}.
We introduce {Learning Impact Measurement (LIM)}, an \textbf{automated} method to evaluate and prioritize training samples based on their alignment with model learning trajectories, enabling efficient resource utilization and scalable implementation. Our method achieves comparable or even superior performance using only \textbf{1,389} samples versus the full \textbf{8,523} samples dataset. 
Notably, while recent data-efficient approaches (e.g., LIMO and s1) show promise with 32B-scale models, we find \textbf{it significantly underperforms at 7B-scale through supervised fine-tuning} (SFT). In contrast, our RL-based LIMR achieves \textbf{16.7\%} higher accuracy on AIME24 and outperforms LIMO and s1 by \textbf{13.0\%} and \textbf{22.2\%} on MATH500. These results fundamentally reshape our understanding of RL scaling in LLMs, demonstrating that precise sample selection, rather than data scale, may be the key to unlocking enhanced reasoning capabilities. For reproducible
research and future innovation, we are open-sourcing LIMR, including implementation of LIM, training and evaluation code, curated datasets, and trained models at \url{https://github.com/GAIR-NLP/LIMR}.

\end{abstract}

\begin{figure}[htbp]
    \centering
    \begin{subfigure}[b]{0.63\textwidth}
        \includegraphics[width=\textwidth]{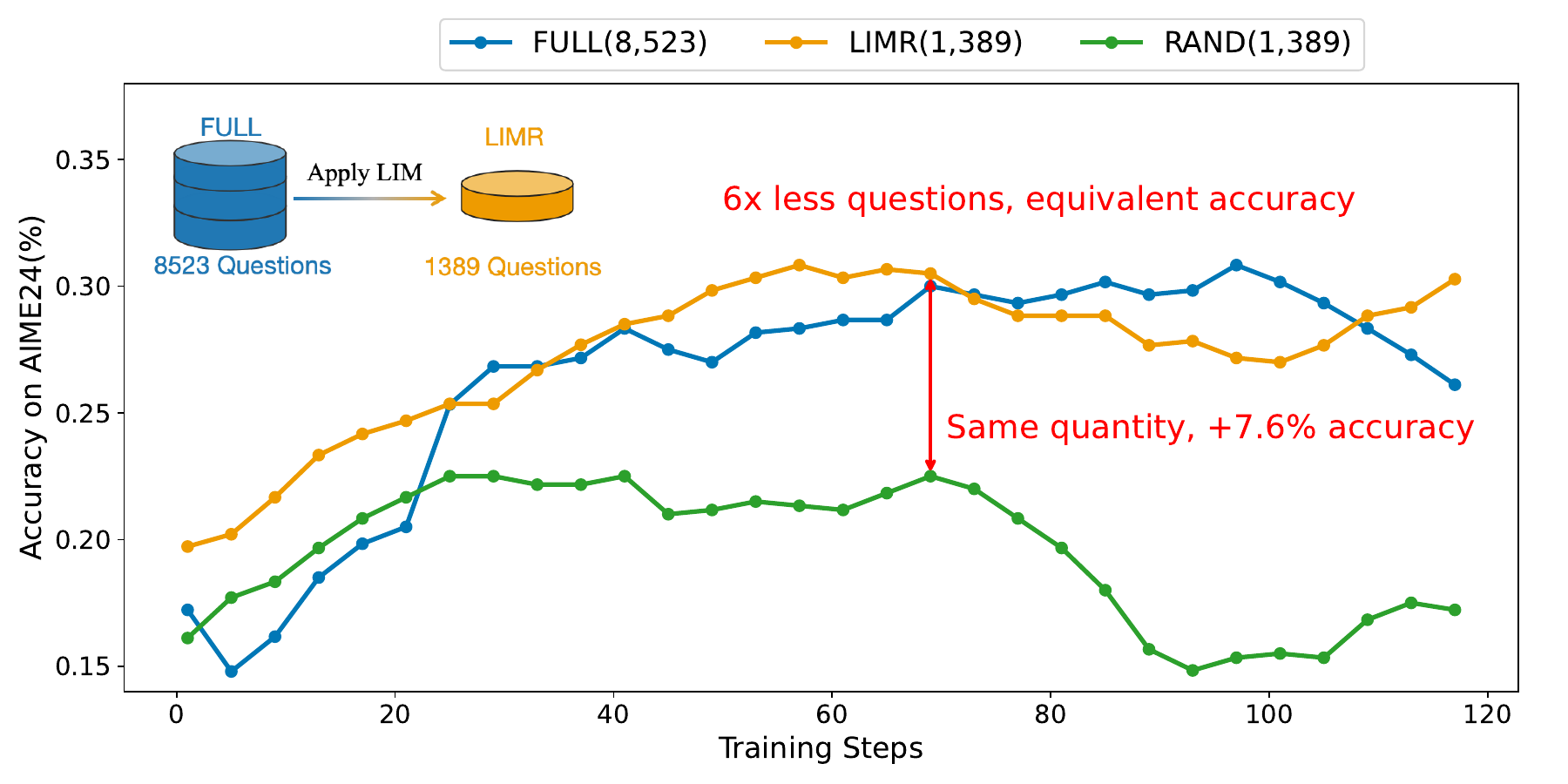} 
        \caption{}
        \label{fig:accuracy_curves}
    \end{subfigure}
    \hfill
    \raisebox{5pt}{
    \begin{subfigure}[b]{0.26\textwidth}
        \includegraphics[width=\textwidth]{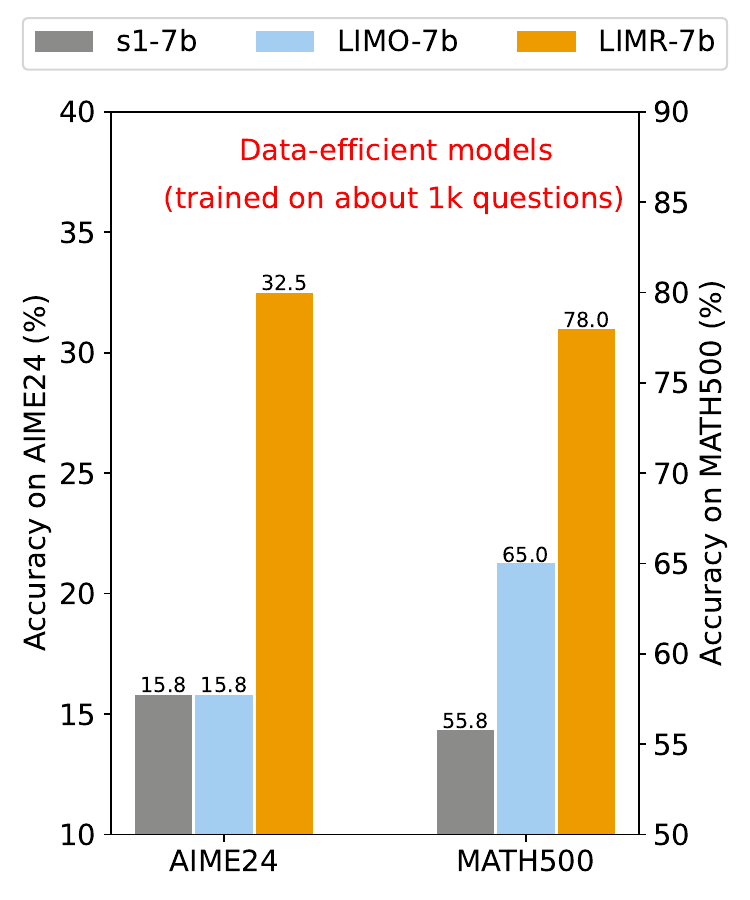}
        \caption{}
        \label{fig:average_accuracy}
    \end{subfigure} }
    \caption{(a) The accuracy on AIME24 across using different training datasets in RL \textbf{without any data distillation and SFT training as cold start.}. Our specifically curated LIMR dataset, a strategically selected subset from the full dataset, MATH (level 3-5), achieved comparable accuracy levels while utilizing less than one-sixth of the data volume. Notably, LIM significantly outperformed a randomly selected dataset of equivalent size, demonstrating the effectiveness of our selective dataset construction methodology. (b) A comparison of different data-efficient models. The results reveal that directly applying SFT on the LIMO~\cite{ye2025limoreasoning} and s1~\cite{muennighoff2025s1simpletesttimescaling} datasets with Qwen-Math-7B yields significantly inferior results compared to using RL with LIMR, implying that, \textbf{for small models, RL is more effective in achieving data efficiency.}}
    
    \label{fig:main-2}
\end{figure}

\newpage

\pagestyle{fancy}
\lhead{\rightmark}
\renewcommand{\headrulewidth}{0.7pt}
\setlength{\headsep}{5mm}

\clearpage

\newpage

\section{Introduction}

Recent advances in Large Language Models (LLMs) have demonstrated the remarkable effectiveness of reinforcement learning (RL) in enhancing complex reasoning capabilities. Models like o1~\cite{openai-o1}, Deepseek R1~\cite{guo2025deepseek}, and Kimi1.5~\cite{kimiteam2025kimik15scalingreinforcement} have shown that RL training can naturally induce sophisticated reasoning behaviors, including self-verification, reflection, and extended chains of thought. However, a critical gap exists in our understanding of RL training: these pioneering works provide \textbf{limited transparency about their training data scale}, making it challenging for the research community to build upon their success. Follow-up open-source efforts (Table~\ref{tab:datasets}) have explored diverse experimental scenarios, from base models to distilled long-form chain-of-thought models, with RL data volumes ranging from 8K~\cite{zeng2025simplerl} to 150K~\cite{cui2025processreinforcementimplicitrewards}, but without clear guidance on optimal data requirements or \textbf{scaling principles}. In this work, we try to explore the scaling dynamics of RL training data by focusing on a foundational scenario: \textbf{starting directly from base models without distillation (similar to the RL scaling setting of Deepseek R1-zero)}.

This lack of understanding of RL training data requirements presents several fundamental challenges:
\begin{itemize}
    \item First, without clear benchmarks for data scale, researchers must rely on trial and error, leading to inefficient resource utilization and potentially suboptimal results. 
    \item Second, the field lacks systematic analysis of how sample quantity impacts model performance, making it difficult to make informed decisions about resource allocation.
\end{itemize}

 More importantly, this uncertainty raises a crucial question: \textbf{Is scaling up RL training data truly the key to improving model performance, or are we overlooking more fundamental factors such as sample quality and selection criteria?}

In this work, we challenge the assumption that larger RL training datasets necessarily lead to better performance. Our key insight is that the quality and relevance of training samples matter far more than their quantity.
Through extensive empirical analysis, we make several surprising observations that fundamentally change our understanding of RL training dynamics:
\begin{enumerate}
    \item We find that a carefully selected subset of RL training samples (1,389) can achieve comparable or even superior performance compared to training with the full dataset (8,523).
    \item Most importantly, we develop an automated quantitative method for evaluating the potential value of RL training samples. Our method, which we call \textbf{Learning Impact Measurement (LIM)}, can effectively predict which samples will contribute most significantly to model improvement. This automated approach eliminates the need for manual sample curation and makes our methodology easily scalable.
    \item Recent approaches like LIMO and s1 have demonstrated the potential of distilled reasoning data efficiency through supervised fine-tuning with 32B models. We find that at 7B-scale, these methods significantly underperform. Our RL-based LIMR achieves 16.7\% higher accuracy on AIME24 (32.5\% vs 15.8\%) and surpasses LIMO and s1 by 13.0\% and 22.2\% on MATH500 (78.0\% vs 65.0\%, 55.8\%), suggesting that RL may be more effective for enhancing reasoning capabilities in data-sparse scenarios.
\end{enumerate}

Our findings have significant implications for the field of LLM development. They suggest that the path to better reasoning capabilities may not lie in simply scaling up RL training data, but rather in being more selective about which samples to use. This insight could dramatically reduce the computational resources required for effective RL training while potentially improving final model performance. Furthermore, our automated sample evaluation method provides a practical tool for researchers and practitioners to implement these insights in their own work. For reproducible research and future innovation, we release all LIMR artifacts openly, including LIMR dataset and model, all training and evaluation code, and implementation details of LIM.

\section{Methodology}

\begin{table}[h]
    \centering
    \small
    \begin{tabular}{lccr}
        \toprule
        Methods & Init Model & Long CoT Distillation &  \#Questions \\
        \midrule
        STILL-3~\cite{Slow_Thinking_with_LLMs_3_Preview} & Instruct & Yes & 29,925 \\
        DeepScaleR~\cite{deepscaler2025} & Instruct & Yes & 40,314 \\
        Sky-T1~\cite{sky_t1_2025} & Instruct & Yes & 45,000 \\
        \midrule
        THUDM-T1~\cite{hou2025advancing} & Instruct & No & 30,000 \\
        PRIME~\cite{cui2025processreinforcementimplicitrewards} & Instruct & No & 150,000 \\
        \midrule
        SimpleRL~\cite{zeng2025simplerl} & Base & No & 8,523 \\
        \midrule
        LIMR & Base & No & 1,389 \\
        \bottomrule
    \end{tabular}
    \caption{Comparison of various methods. The ``init model" refers to the type of the initial actor model, we performance RL directly on the base model. ``Long CoT Distillation" indicates whether the initial model distills long CoT for cold start.}
    \label{tab:datasets}
\end{table}

We present Learning Impact Measurement (LIM), a systematic approach to quantify and optimize the value of training data in reinforcement learning. Our method addresses the critical challenge of data efficiency in RL training by analyzing learning dynamics to identify the most effective training samples.

\subsection{Learning Dynamics in RL Training}

\begin{figure}[htbp]
    \centering
    \begin{subfigure}[b]{0.45\textwidth}
        \includegraphics[width=\textwidth]{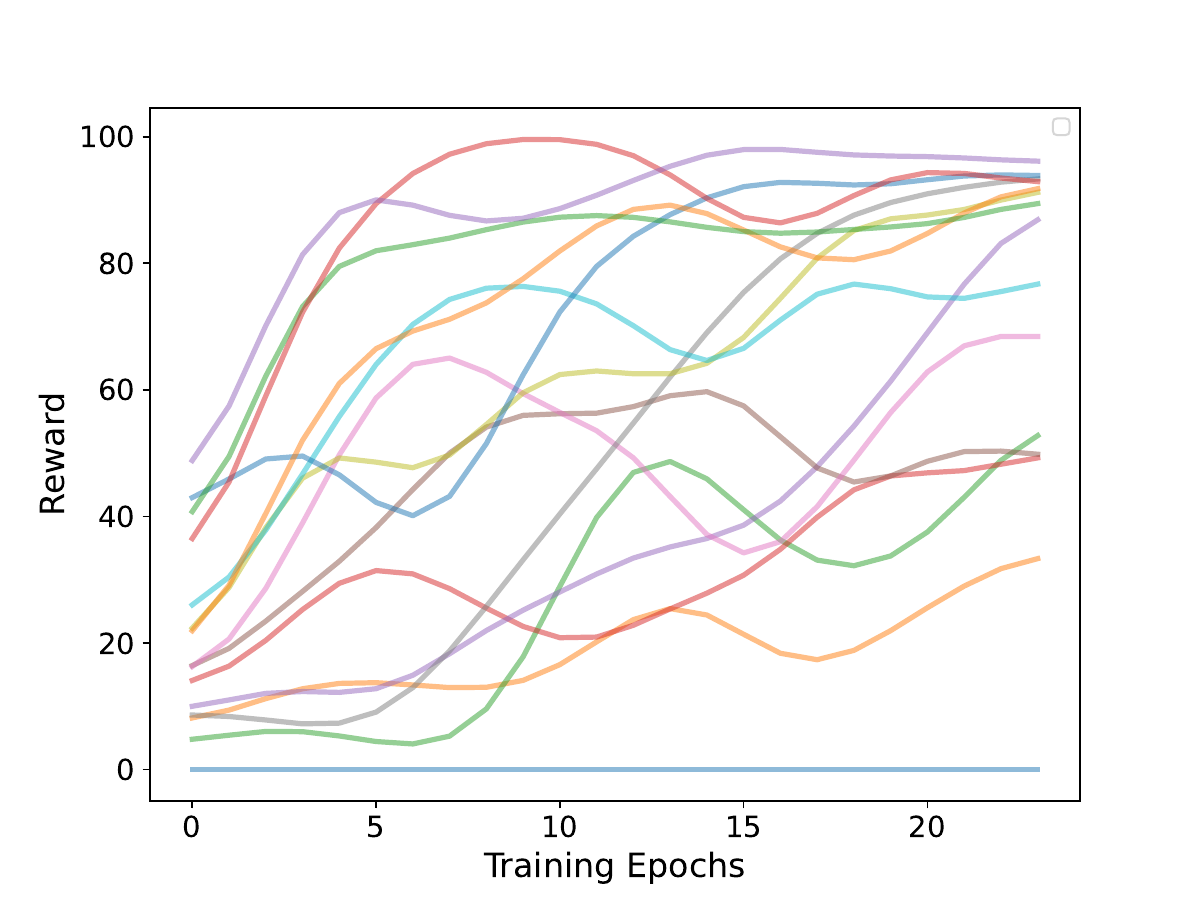} 
        \caption{}
        \label{fig:accuracy_curves}
    \end{subfigure}
    \hspace{0.3cm}
    \begin{subfigure}[b]{0.45\textwidth}
        \includegraphics[width=\textwidth]{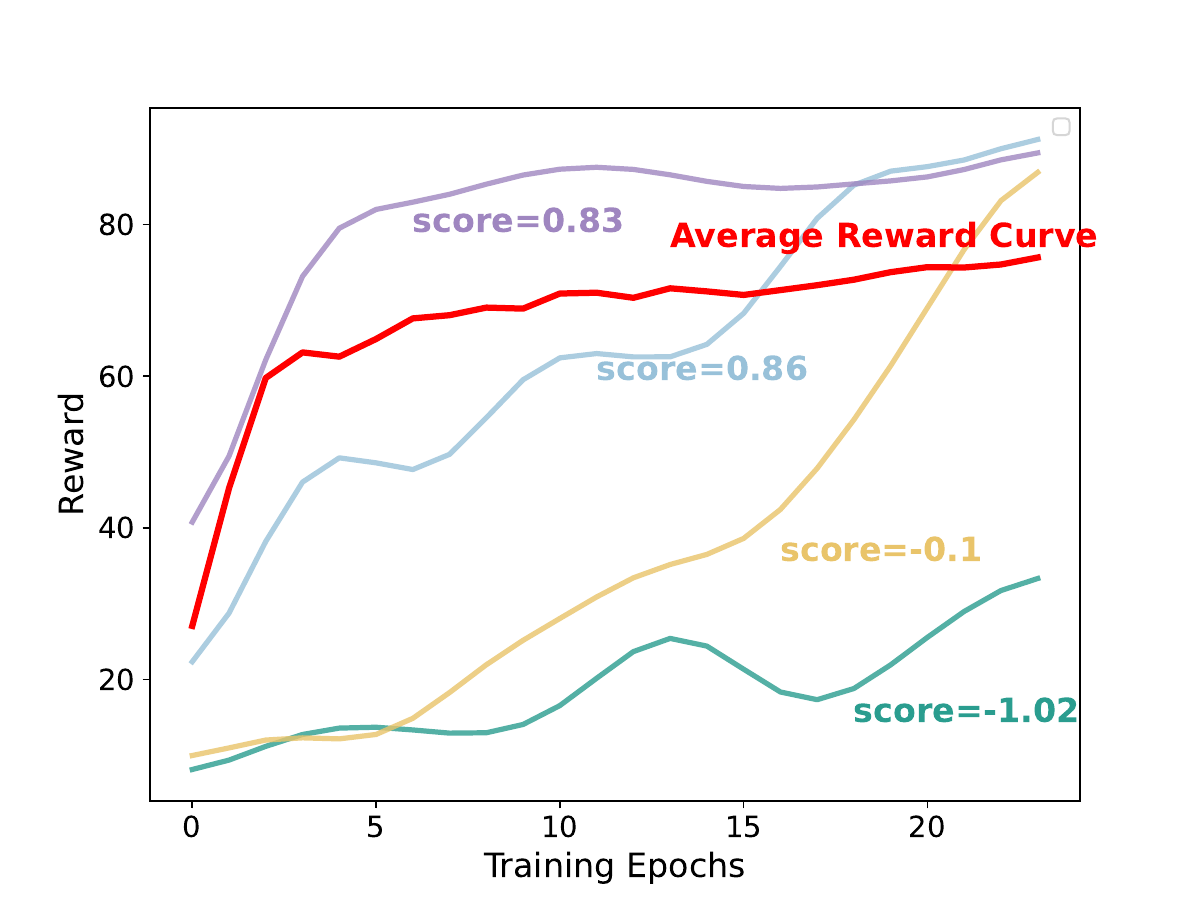}
        \caption{}
        \label{fig:average_accuracy}
    \end{subfigure}
    \caption{(a) Learning dynamics analysis of training samples from MATH-FULL dataset across epochs. Solution reward trajectories reveal diverse patterns: samples maintaining near-zero rewards, samples quickly achieving high rewards, and those showing dynamic learning progress with varying improvement rates. (b) Sample learning trajectories compared against the average reward curve (red). Higher LIM scores reflect better alignment with model's learning trajectory, where trajectories showing similar growth patterns receive higher scores.}
    \label{fig:main-2}
\end{figure}



To understand the relationship between training data and model improvement, we conducted extensive analysis using the MATH-FULL dataset~\cite{hendrycks2021measuring}, which contains 8,523 mathematical problems of varying difficulty levels (3-5). Our investigation reveals that different training samples contribute unequally to model learning, contrary to the conventional approach of treating all samples uniformly. As illustrated in Figure~\ref{fig:accuracy_curves}, we observe diverse learning trajectories: some samples exhibit stable performance patterns, while others show complex learning dynamics that appear to drive significant model improvements.

These observations lead to our key insight: the value of training data in RL can be systematically measured by examining how well individual samples align with the model's overall learning progression. This understanding forms the foundation of LIM, our proposed method for quantifying sample effectiveness.

\subsection{Learning Impact Measurement (LIM)}
\label{subsec:lim}

LIM centers on a model-aligned trajectory analysis that evaluates training samples based on their contribution to model learning. Our key finding is that samples whose learning patterns complement the model's overall performance trajectory tend to be more valuable for optimization.

\subsubsection{Model-aligned Trajectory Analysis}

Given that neural network learning typically follows a logarithmic growth pattern, we use the model's average reward curve as a reference for measuring sample effectiveness (Figure~\ref{fig:average_accuracy}):

$$r_{\text{avg}}^k = \frac{1}{N} \sum_{i=1}^{N} r_i^k, k=1, ..., K$$

where $r_i^k$ represents the reward of sample $i$ at epoch $k$, and $N$ is the total number of samples.

For each sample, LIM computes a normalized alignment score:

$$\text{s}_i = 1 - \frac{\sum_{k=1}^{K} (r_i^k - r_{\text{avg}}^k)^2}{\sum_{k=1}^{K} (1 - r_{\text{avg}}^k)^2}, i=1, ..., N$$

This score quantifies how well a sample's learning pattern aligns with the model's overall learning trajectory, with higher scores indicating better alignment.

\subsubsection{Sample Selection Strategy}
Based on the alignment scores, LIM implements a selective sampling strategy:
$s_i > \theta$ where $\theta$ serves as a quality threshold that can be adjusted according to specific requirements. In our experiments, setting $\theta = 0.6$ yielded an optimized dataset (LIMR) of 1,389 high-value samples from the original dataset.

\subsection{Baseline Data Selection Methods}

While developing our core methodology, we explored several alternative approaches that helped inform and validate our final method. These approaches, provide valuable insights into data selection in RL.

\paragraph{Random Sampling baseline (RAND)} randomly selects 1,389 samples from MATH-FULL to match the size of our main approach, providing a fundamental reference point for evaluating selective sampling effectiveness.


\paragraph{Linear Progress Analysis method (LINEAR)} evaluates samples based on their consistency in showing steady improvements across training epochs. While this approach captures samples with gradual progress, it often misses valuable samples that show rapid early gains followed by stabilization. Using a threshold of $\theta = 0.7$, this method yields 1,189 samples.

\subsection{Reward Design}
Similar to deepseek r1~\cite{guo2025deepseek}, we use a rule-based reward function. Specifically, for a correct answer, the reward is 1; for an incorrect but properly formatted answer, the reward is -0.5; and for a answer with formatting errors, the reward is -1. Formally, this can be expressed as: 
$$
R(\text{answer}) =
\begin{cases} 
1 & \text{if the answer is correct,} \\
-0.5 & \text{if the answer is incorrect but well-formatted,} \\
-1 & \text{if the answer has formatting errors.}
\end{cases}
$$

\section{Experiment}
\subsection{Experimental Setup}
\paragraph{Training}
We conduct RL training using PPO~\cite{schulman2017proximalpolicyoptimizationalgorithms} algorithm implemented in the OpenRLHF~\cite{hu2024openrlhf} framework. Using Qwen2.5-Math-7B~\cite{yang2024qwen25mathtechnicalreportmathematical} as our initial policy model, we configure the rollout batch size as 1,024 and generate 8 samples per prompt with a temperature of 1.2 during exploration. The training process uses a batch size of 256, with learning rates set to 5e-7 and 9e-6 for the actor and critic models respectively, and a KL coefficient of 0.01.

%
\paragraph{Evaluation}
We conducted experimental evaluations on multiple challenging benchmarks, including: (1) MATH500; (2) AIME2024; (4) AMC2023. To accelerate the evaluation process, we utilized the vLLM~\cite{kwon2023efficient} framework. For AIME24, AMC23, due to the limited number of questions (30 and 40 respectively), we performed 4 sampling runs per question with a temperature of 0.4. For MATH500, we employed greedy decoding for inference.

\definecolor{lightblue}{RGB}{173, 216, 230}
\subsection{Main Results}
\begin{table}[h]
\centering
\caption{Main results on difficult math benchmarks.}
\label{table:main}
\resizebox{0.8\textwidth}{!}{
\begin{tabular}{llcccc}
\toprule
\textbf{Method} & \textbf{\#Questions} & \textbf{AIME2024} & \textbf{MATH500} & \textbf{AMC2023} & \textbf{AVG.} \\
\midrule
Qwen-Math-7B & - & 16.7 & 52.4 & 52.5 & 40.5\\
\midrule
Qwen-Math-7B-FULL & 8,523 & 32.5 & 76.6 & 61.9 & 57.0 \\
Qwen-Math-7B-RAND & 1,389 & 25.8 & 66.0 & 56.3 & 49.4 \\
Qwen-Math-7B-LINEAR & 1,138 & 28.3 & 74.6 & 61.9 & 54.9 \\
\midrule
\rowcolor{lightblue}
LIMR & 1,389 & \textbf{32.5} & \textbf{78.0} & \textbf{63.8} & \textbf{58.1} \\
\bottomrule
\end{tabular}}
\end{table}

\begin{figure}[htbp]
    \centering
    \begin{subfigure}[b]{0.31\textwidth}
        \includegraphics[width=\textwidth]{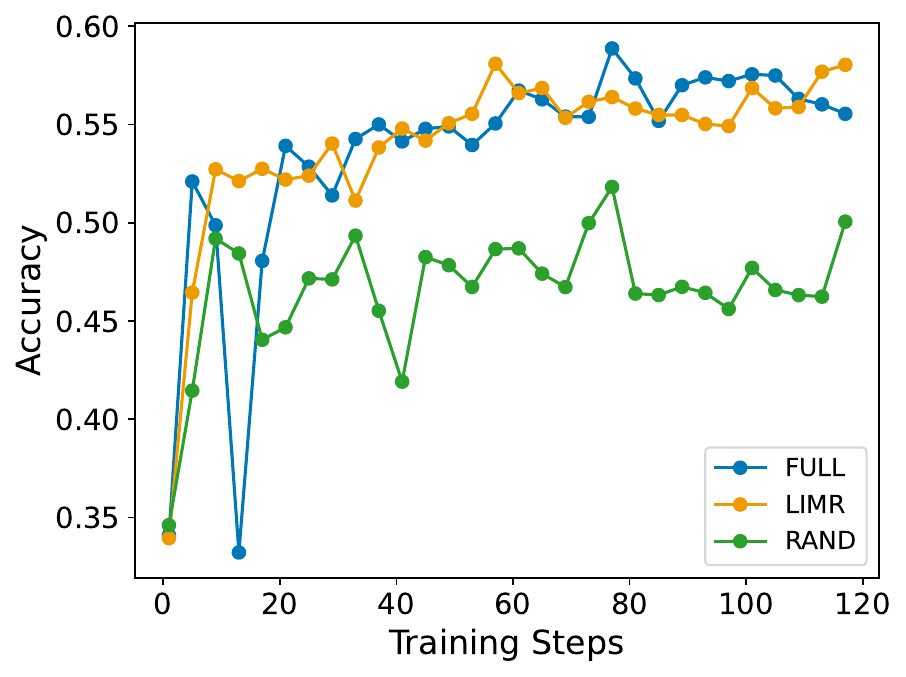} 
        \caption{Average Accuracy on Benchmarks}
        \label{fig:main-1-1}
    \end{subfigure}
    \hfill 
    \begin{subfigure}[b]{0.31\textwidth}
        \includegraphics[width=\textwidth]{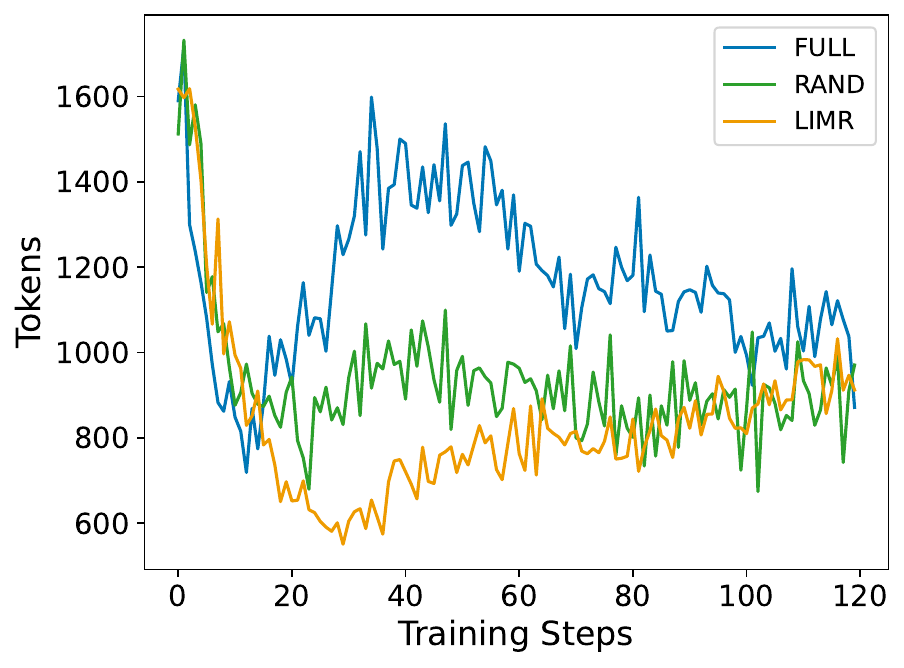} 
        \caption{Response Length}
        \label{fig:main-1-2}
    \end{subfigure}
    \hfill 
    \begin{subfigure}[b]{0.31\textwidth}
        \includegraphics[width=\textwidth]{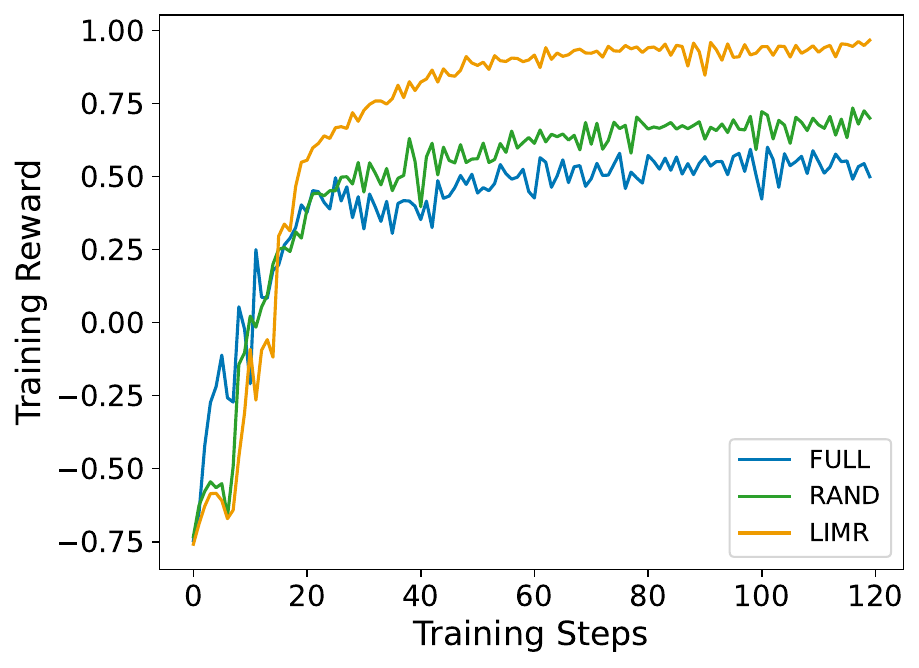} 
        \caption{Training Reward}
        \label{fig:main-1-3}
    \end{subfigure}
    \caption{Performance and training dynamics}
    \label{fig:main-1}
\end{figure}

As illustrated in Table~\ref{table:main}, directly applying RL to Qwen-Math-7B using the MATH-FULL dataset resulted in a significant performance improvement. Different data selection strategies, however, led to notable variations in performance. Training with the MATH-RAND dataset results in an average accuracy drop of 8.1\% compared to using the full dataset, whereas MATH-LINEAR incurs only a ~2\% loss. More notably, LIMR, despite an 80\% reduction in dataset size, performs nearly on par with MATH-FULL. This further supports the notion that in RL, only a small subset of questions plays a critical role.  

Additionally, we analyze the evolution of various metrics during RL training on the MATH-FULL, MATH-RAND, and LIMR datasets. As shown in Figur~\ref{fig:main-1-1}, the accuracy curves of LIMR and MATH-FULL are nearly identical, both significantly outperforming MATH-RAND. Meanwhile, Figure~\ref{fig:main-1-2} indicates that the training curve for MATH-FULL exhibits instability in terms of sequence length, whereas the corresponding curve for LIMR initially declines before gradually increasing. Figure~\ref{fig:main-1-3} further illustrates differences in training rewards: the reward curve for LIMR rises more rapidly and ultimately approaches 1.0. This suggests that during RL, the model effectively utilizes the LIMR dataset for learning.

Figure~\ref{fig:main-2} presents a comparative analysis of model performance across three challenging benchmarks. The results demonstrate that LIMR achieves performance comparable to MATH-FULL on all three benchmarks, while significantly outperforming the RAND baseline. Notably, LIMR's consistent excellence on both AIME24 and AMC23 datasets provides compelling evidence that its enhanced performance is not attributable to overfitting to a single dataset, but rather reflects a genuine improvement in the model's mathematical reasoning capabilities compared to the RAND.

\begin{figure}[htbp]
    \centering
    \begin{subfigure}[b]{0.31\textwidth}
        \includegraphics[width=\textwidth]{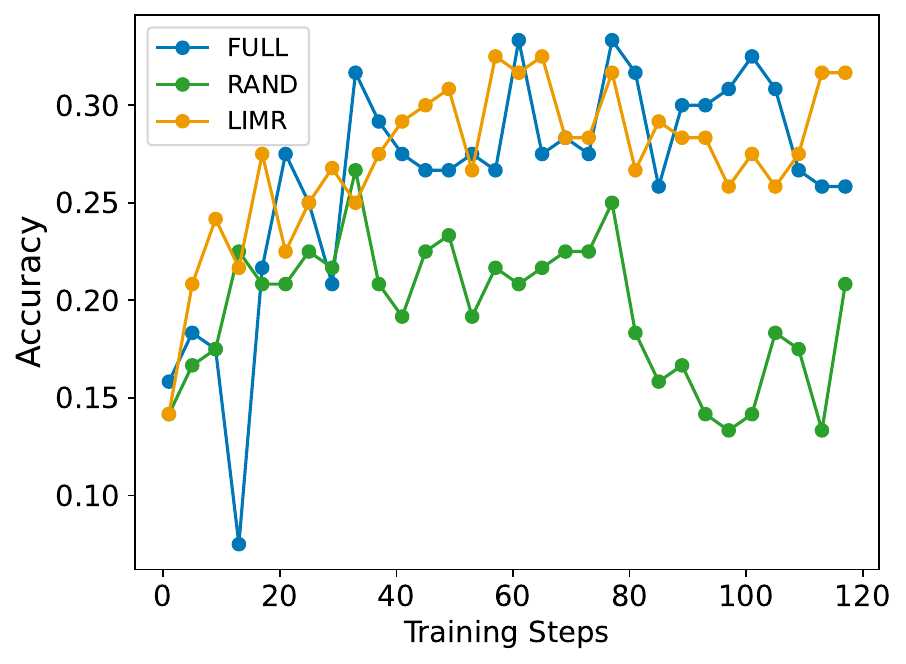} 
        \caption{Accuracy on AIME24}
        \label{fig:main-2-1}
    \end{subfigure}
    \hfill 
    \begin{subfigure}[b]{0.31\textwidth}
        \includegraphics[width=\textwidth]{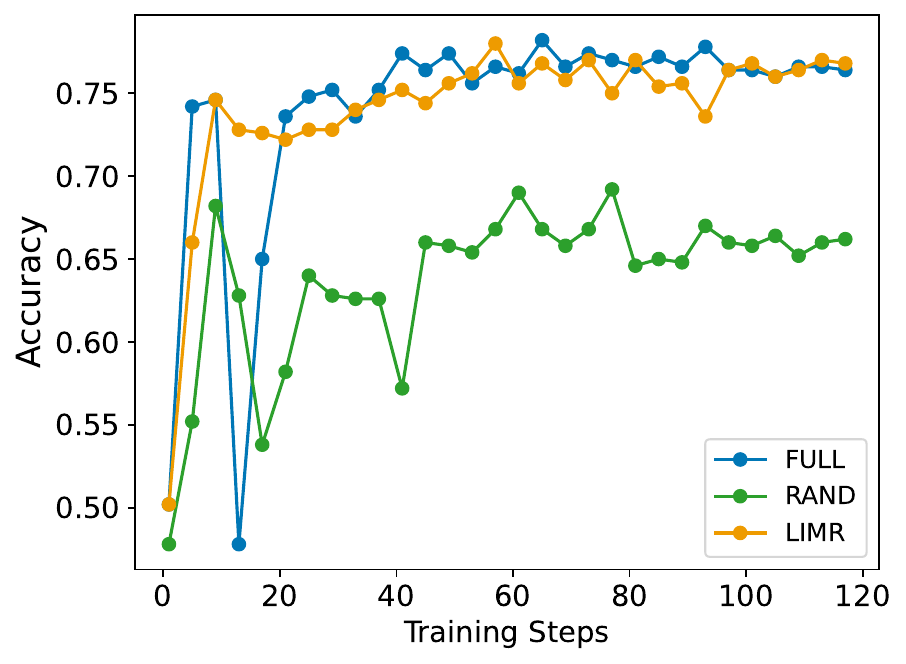} 
        \caption{Accuracy on MATH500}
        \label{fig:main-2-2}
    \end{subfigure}
    \hfill 
    \begin{subfigure}[b]{0.31\textwidth}
        \includegraphics[width=\textwidth]{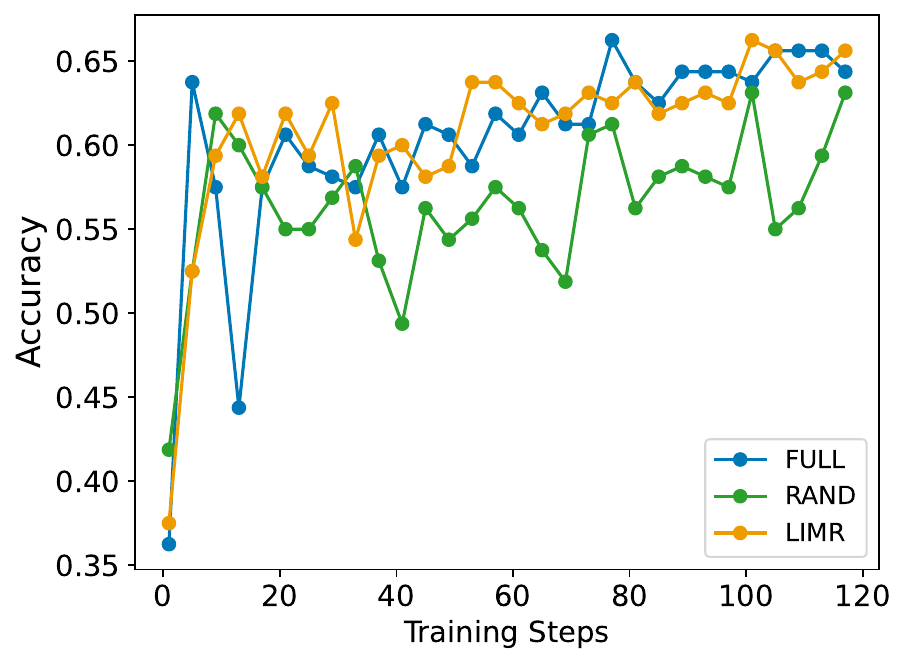} 
        \caption{Accuracy on AMC23}
        \label{fig:main-2-3}
    \end{subfigure}
    \caption{Accuracy on various benchmarks}
    \label{fig:main-2}
\end{figure}

\subsection{RL Outperforms SFT in Data Efficiency}

\begin{table}[h]
\centering
\caption{Performance difference of three data efficient models.}
\label{table:baseline}
\resizebox{0.75\textwidth}{!}{
\begin{tabular}{llcccc}
\toprule
\textbf{Method} & \textbf{\#Questions} & \textbf{AIME2024} & \textbf{MATH500} & \textbf{AMC2023} & \textbf{AVG.}\\
\midrule
Qwen-Math-7B & - & 16.7 & 52.4 & 52.5 & 40.5 \\
\midrule
Qwen-Math-7B-s1 & 1,000 & 15.8 & 55.8 & 42.5 & 38.0 \\
Qwen-Math-7B-LIMO & 817 & 15.8 & 65.0 & 56.3 & 45.7 \\
\midrule
\rowcolor{lightblue}
LIMR & 1,389 & \textbf{32.5} & \textbf{78.0} & \textbf{63.8} & \textbf{58.1} \\
\bottomrule
\end{tabular}
}
\end{table}

Both LIMO~\cite{ye2025limoreasoning} and s1~\cite{muennighoff2025s1simpletesttimescaling} emphasize that only a small amount of data is needed to unlock the reasoning potential of models. However, we found that in scenarios with limited data and small models (e.g., 7B models), using reinforcement learning (RL) is more effective than distilling data from larger models and performing imitation learning.

Specifically, we fine-tuned Qwen-2.5-Math-7B using 1,000 pieces of data from s1 and 817 pieces of data from LIMO via supervised fine-tuning and compared it with LIMR. The experimental results show that, with the same 1k questions, Compared to LIMO and s1, LIMR has achieved a relative improvement of over 100\% on AIME, and at least a 10\% accuracy increase on AMC23 and MATH500. This further underscores the importance of selecting data that is suitable for the model rather than blindly opting for more challenging data.

\section{Conclusion}

In this work, we challenge the conventional wisdom that scaling up RL training data is necessary for improving LLM reasoning capabilities. Through the introduction of Learning Impact Measurement (LIM), we demonstrate that a carefully selected subset of 1,389 samples can match or exceed the performance of the full 8,523-sample dataset across multiple challenging mathematical benchmarks. Our automated LIM methodology not only provides a practical, scalable solution for researchers to implement efficient RL training but also reveals that the path to enhanced reasoning capabilities may lie in optimizing sample quality rather than increasing data quantity. Additionally, our comparison with supervised fine-tuning approaches demonstrates that RL, when combined with efficient data selection, can be particularly effective for smaller models with limited data, suggesting potential applications of our methodology beyond mathematical reasoning to other domains where RL is applied in language models.

\newpage

\bibliographystyle{acl_natbib}
\bibliography{related}

\end{document}